# PersianLLaMA: Towards Building First Persian Large Language Model

Mohammad Amin Abbasi[1], Arash Ghafouri[2*], Mahdi Firouzmandi[2], Hassan Naderi[1] and Behrouz Minaei Bidgoli[1]

[1] Department of Computer Engineering Iran University of Science and Technology, Tehran, Iran.

[2] Department of Artificial Intelligence and Cognitive Science, Imam Hossein Comprehensive University, Tehran, Iran.

* Corresponding author. E-mail: krghafouri@ihu.ac.ir;

Contributing authors E-mail: m_abbasi1378@comp.iust.ac.ir; firouzmandi@ihu.ac.ir; Naderi@iust.ac.ir; B_minaei@iust.ac.ir.

**ABSTARCT**

Despite the widespread use of the Persian language by millions globally, limited efforts have been made in natural language processing for this language. The use of large language models as effective tools in various natural language processing tasks typically requires extensive textual data and robust hardware resources. Consequently, the scarcity of Persian textual data and the unavailability of powerful hardware resources have hindered the development of large language models for Persian. This paper introduces the first large Persian language model, named PersianLLaMA, trained on a collection of Persian texts and datasets. This foundational model comes in two versions, with 7 and 13 billion parameters, trained on formal and colloquial Persian texts using two different approaches. PersianLLaMA has been evaluated for natural language generation tasks based on the latest evaluation methods, namely using larger language models, and for natural language understanding tasks based on automated machine metrics. The results indicate that PersianLLaMA significantly outperforms its competitors in both understanding and generating Persian text. PersianLLaMA marks an important step in the development of Persian natural language processing and can be a valuable resource for the Persian-speaking community. This large language model can be used for various natural language processing tasks, especially text generation like chatbots, question-answering, machine translation, and text summarization.

**Keywords:** Large language models, Persian language, Natural language generation, Natural language understanding.

## 1 INTRODUCTION

In recent decades, significant advances in artificial intelligence have led to the emergence of large language models as pivotal and transformative tools in the field of artificial intelligence and natural language processing. These AI models, utilizing deep neural networks and learning from extensive datasets, are capable of understanding and generating texts with structures similar to human language.

Language modeling is a primary approach for advancing machine language intelligence. Generally, the aim of language modeling is to model the probability of word sequences to predict the likelihood of future words. Research in language modeling has gained extensive attention in recent years, leading to the emergence of pre-trained language models. Initially, the ELMo(Peters et al., 2018) model was introduced to understand content in word embedding, which involved pre-training a bi-directional LSTM (biLSTM)(Staudemeyer & Morris, 2019) network and then fine-tuning the biLSTM network based on sub-tasks. Also, the BERT(Devlin et al., 2018) model, utilizing the structure of Transformers and the Encoder part of the Transformer model, was introduced. Pre-trained on large unlabelled text datasets, this model was ready for use in sub-tasks, significantly enhancing the performance of natural language understanding tasks. These models often require the pre-trained language model to be finely tuned for compatibility with sub-tasks.

Subsequently, researchers discovered that increasing the number of parameters in pre-trained language models generally leads to better results in downstream tasks. Some studies in this field have explored the performance of increasingly large pre-trained language models, such as the GPT-3(Floridi & Chiriatti, 2020) model with 175 billion parameters and the PaLM(Chowdhery et al., 2023) model with 540 billion parameters. Thus, the research community coined the term "large language models" for these types of models. A notable practical application of large language models is the ChatGPT(Ray, 2023) model, which uses the GPT series models for conversation, demonstrating remarkable ability in interactions with humans.

LLaMA(Touvron, Lavril, et al., 2023; Touvron, Martin, et al., 2023) is a large language model developed by Facebook. It is an artificial neural network that has been trained on a vast dataset of text and code. LLaMA comprises a collection of open-source language models and is available in two versions. The first and second versions have parameters ranging from 7 to 65 billion and from 7 to 70 billion, respectively, and are competitive with state-of-the-art language models. Notably, LLaMA-13B, a version of LLaMA with 13 billion parameters, performs better in a multitude of tasks compared to GPT-3, while being approximately ten times smaller in size. Unlike most large language models, LLaMA has been trained on specific and open-source datasets and has achieved state-of-the-art performance without relying on proprietary data. Additionally, fine-tuning LLaMA on different data sets leads to very good results.

However, large language models also face challenges and limitations; due to their numerous parameters, these models require significant computational resources for training and usage. Additionally, concerns about their capability to generate incorrect content or false news exist.

Despite the remarkable capabilities of these large language models, they have been predominantly developed in English and limiting their applicability to other languages. This language bias has created a significant gap in the development and utilization of language models for non-English languages, including Persian. The Persian language, spoken by millions worldwide, lacks comprehensive language models that can effectively understand and generate Persian text.

Recognizing this gap, our research aims to develop PersianLLaMA, the first large language model dedicated to the Persian language. The development of this model is not only a technical challenge but also a step towards linguistic inclusivity in the field of AI and natural language processing. By focusing on the Persian language, PersianLLaMA seeks to provide a robust tool for understanding and generating Persian text, which can be utilized in various applications, from chatbots to text summarization.

## 2 RELATED WORKS

Given the absence of a large language model in Persian, this section aims to examine the closest works related to our research. These include large Asian language models based on the LLaMA framework and both single-language and multilingual pre-trained models that support Persian.

### 2.1 LLaMA-Based Models

Several studies have recently been conducted on developing large language models for Asian languages. For instance, TAMIL-LLaMA(Balachandran, 2023), an Asian language model that leverages LLaMA and incorporates 16,000 Tamil tokens, has shown significant improvements. This model employs the LoRA(Hu et al., 2021) technique for efficient training on Tamil datasets, including translated Alpaca datasets and a custom subset of the OpenOrca(Mukherjee et al., 2023) dataset for fine-tuning. Performance evaluations demonstrate notable enhancements in Tamil text generation, with the 13B version outperforming OpenAI's GPT-3.5-turbo in Tamil language tasks.

SeaLLMs(Nguyen et al., 2023) is a suite of language models dedicated to Southeast Asian (SEA) languages, built on the LLaMA-2 model with extended vocabularies and specific configurations to cater to the cultural nuances of SEA languages. These models excel in various linguistic tasks and comply with local cultural standards and legal considerations. They perform better than models like ChatGPT-3.5 in non-Latin SEA languages, including Thai, Khmer, Lao, and Burmese. Additionally, SeaLLMs have significantly advanced AI tools with cultural awareness.

TAIWAN-LLM(Lin & Chen, 2023) is a large language model for Traditional Chinese, particularly used in Taiwan. Available in 7 and 13 billion parameter versions, it is developed using a comprehensive pre-training dataset and fine-tuning datasets that capture the linguistic and cultural essence of Taiwan. TAIWAN-LLM outperforms many models in understanding and generating Traditional Chinese text.

### 2.2 Pre-Trained Persian Language Models

Persian language models play a crucial role in advancing research related to Persian language processing. These models, pre-trained on extensive Persian text datasets, are adept at accurately comprehending and interpreting Persian language structures and meanings.

AriaBERT(Ghafouri et al., 2023) is recognized as the leading language model in understanding the Persian language. Utilizing the RoBERTa(Liu et al., 2019) architecture and the Byte-Pair Encoding(Sennrich et al., 2015) tokenizer, AriaBERT has been trained with over 32 gigabytes of various Persian textual data. This includes a mix of colloquial and formal texts comprising tweets, news, poetry, medical texts, encyclopedia articles, user comments on websites, and other text types.

ParsBERT is a monolingual model based on Google's BERT architecture, trained on a large corpus of Persian texts covering diverse topics with more than 3.9 million documents and 16 gigabytes of data. It outperforms multilingual models like Multilingual BERT (M-BERT) in various tasks such as sentiment analysis, text classification, named entity recognition, and named entity disambiguation in Persian texts.

SINA-BERT(Taghizadeh et al., 2021) Introduced as a BERT-based language model, is published for high-quality Persian language coverage in the medical field. It utilizes an extensive collection of medical content, including both formal and informal texts. The primary use of this model includes medical question classification, sentiment analysis in medicine, and medical question retrieval.

The mT5(Xue et al., 2020) model, known for its text-to-text transformation, is another prominent NLP model developed by Google. Based on the Transformer architecture, it frames all NLP tasks as converting input text to output text. This versatile adaptability is evident during fine-tuning, where the model aligns with specific tasks using labeled data. The multilingual version, MT5, trained on a vast corpus of web-crawled text in 101 languages including Persian, exemplifies this adaptability.

The mGPT(Shliazhko et al., 2022) models, similar to GPT2(Radford et al., 2019), come in 1.3 and 13 billion parameter versions and are trained on 60 languages. Their training dataset includes mc4(Raffel et al., 2020) and Wikipedia, totaling approximately 600 gigabytes. mGPT, built on the decoder component of Transformer architecture, has been evaluated in various scenarios, including language modeling, downstream task evaluation, and knowledge-based searches, showing remarkable performance in Persian among other supported languages. Table 1 shows a comprehensive overview of recent language models and Figure 1 shows a timeline of existing large language models in recent years.

Table 1: Comprehensive overview of recent language models

| Model | Release Time | Parameter Size | Base Model | Train Data Size | Hardware | Type |
|---|---|---|---|---|---|---|
| Gemini(Anil et al., 2023) | Dec-2023 | 176B | - | 6.3T tokens | - | Causal decoder |
| SeaLLMs | Dec-2023 | 13B | LLaMA2 | - | - | Causal decoder |
| TAMIL-LLaMA | Nov-2023 | 13B | LLaMA2 | - | - | Causal decoder |
| TAIWAN-LLM | Nov-2023 | 13B | LLaMA2 | 35B tokens | - | Causal decoder |
| AriaBERT | Nov -2023 | 355M | RoBERTa | 102M documents | 4 A100 40G | Masked LM |
| Falcon-40B(Penedo et al., 2023) | Jul-2023 | 40B | - | 1T tokens | 384*A100 40GB | Causal decoder |
| LLaMA 2 | Jul-2023 | 70B | - | 2T tokens | 2000*80G A100 | Causal decoder |
| Vicuna(Zheng et al., 2023) | Jun-2023 | 13B | LLaMA | 125K instructions | - | auto-regressive language model |
| PaLM2 | May-2023 | 16B | - | 100B tokens | - | Causal decoder |
| Alpaca | May-2023 | 13B | LLaMA | 52K instructions | - | - |
| WizardLM(Xu et al., 2023) | Apr-2023 | 7B | LLaMA | 250k instructions | - | - |
| Koala | Apr-2023 | 13B | LLaMA | 472 instructions | 8*A100 | - |
| umT5(Chung et al., 2023) | Apr-2023 | 13B | T5 | 29T characters | - | Encoder-decoder |
| GPT-4(Koubaa, 2023) | Mar-2023 | 1.8T | - | - | - | Causal decoder |
| LLaMA | Feb-2023 | 65B | - | 1.4T tokens | 2048*A100 80G | Causal decoder |
| mGPT | Apr-2022 | 13B | GPT-3 | 440B tokens | 256*Nvidia V100 | Causal decoder |
| PaLM | Apr-2022 | 540B | - | 780B tokens | 6144 TPU v4 | Causal decoder |
| SINA-BERT | Apr-2021 | 110M | BERT | 2.8M documents | - | Masked LM |
| mT5 | Oct-2020 | 13B | T5 | 1T tokens | - | Encoder-decoder |
| ParsBERT | May-2020 | 110M | BERT | 3.9M documents | - | Masked LM |
| GPT-3 | May-2020 | 175B | - | 300B tokens | - | Causal decoder |

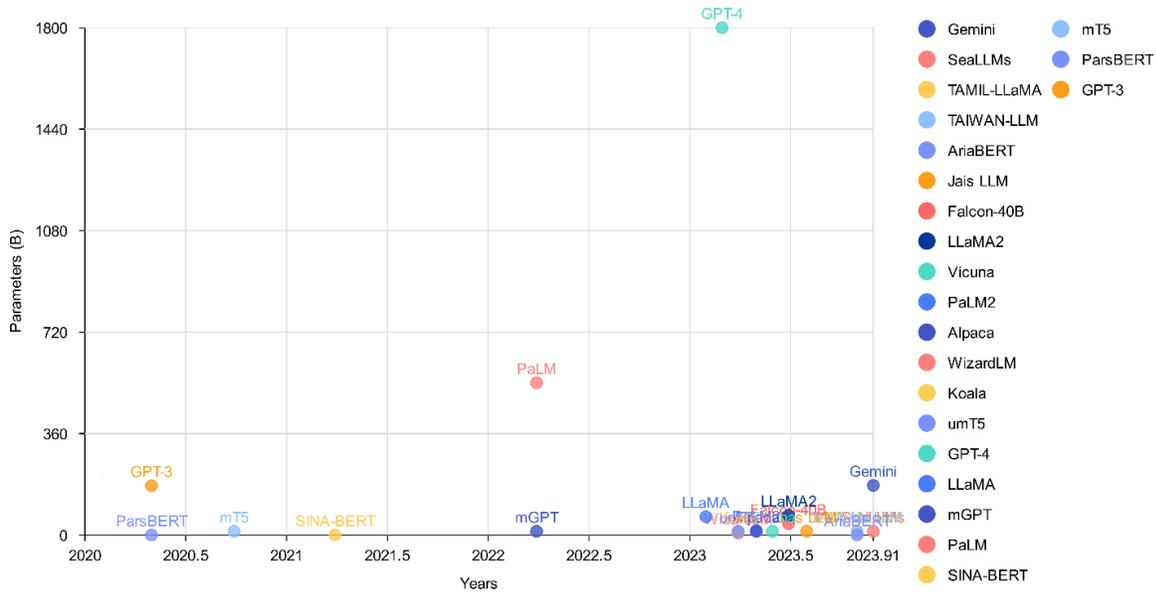

Figure 1: A timeline of existing large language models in recent years. The timeline was established mainly according to the release date (e.g., the submission date to arXiv) of the technical paper for a model

## 3 PROPOSED METHOD

In this research, two distinct approach for developing a large Persian language model are introduced and compared. The first approach involves training a LLaMA 2 model from scratch using a Persian text dataset exceeding 90 gigabytes. A Persian tokenizer was trained using this dataset. Due to hardware resource limitations, a 7 billion parameter LLaMA 2 model was utilized, and the entire neural network was trained from scratch. The second approach involves training a LLaMA model as an adapter on a pre-trained English LLaMA 2 model. Initially, a Persian tokenizer was trained on Persian Wikipedia texts, then combined with the tokenizer of the English LLaMA 2 model. Subsequently, an adapter neural network was placed on the English LLaMA 2 model, and its weights were trained. The 13 billion parameter LLaMA 2 model was used in this approach, as it requires fewer hardware resources, only necessitating training of the adapter part while keeping the core neural network unchanged. Details of both approachs are thoroughly explained in the subsequent sections.

### 3.1 LLaMA-Based Models

To train PersianLLaMA, we selected two datasets comprising various and diverse texts to ensure comprehensive training across different topics. The training dataset includes:

- OSCAR Dataset (Open Super-large Crawled Aggregated Resource)(Suárez et al., 2020): A rich source of multilingual texts collected from the web, widely used in natural language processing and machine learning research. It includes website texts in multiple languages, offering a broad variety of topics and linguistic structures. The Persian section contains 23 million texts (93 GB, 9 billion tokens).

- Persian Wikipedia: A vast and diverse source of textual information, Persian Wikipedia serves as a valuable resource due to its extensive coverage of topics across various fields. It provides a rich and comprehensive linguistic collection for training our model. The dataset allows our model to develop an extensive vocabulary and a precise understanding of Persian language structures and semantics. Utilizing Persian Wikipedia as a foundational dataset ensures that the model is well-prepared to generate coherent and relevant Persian text, covering a wide range of domains. This dataset includes 2.4 million texts (1.3 GB, 184 million tokens).

**3.2 Preprocessing**

In developing the PersianLLaMA text generation model, effective preprocessing of training data is crucial to ensure the quality and coherence of the generated text. This section explains the preprocessing steps undertaken to prepare the training dataset for PersianLLaMA, utilizing a combination of techniques and libraries suited for Persian text.

The process begins with text normalization using the Hazm library, a powerful tool for processing Persian texts. This step involves correcting common issues related to spacing and normalizing characters in Persian texts. Subsequently, HTML tags present in texts are removed. Additional tasks include managing Unicode, normalizing spaces, and removing links, emails, and phone numbers. Persian texts may contain various punctuation marks that can impact analysis and underlying tasks. To address this, superfluous punctuation marks are removed from the texts. Additionally, mentions starting with '@' in tweets are deleted. Finally, multiple consecutive spaces are condensed into a single space. In summary, text preprocessing meticulously cleans and normalizes the training dataset, ensuring its suitability for training the Persian text generation model.

**3.3 Tokenizer**

In the development of natural language processing models for Persian, training an appropriate tokenizer is a fundamental and vital aspect. The tokenizer divides text into smaller units (tokens) to make it understandable and usable by neural models. PersianLLaMA models use the SentencePiece tokenizer, an unsupervised text tokenizer used for neural network-based text generation systems. SentencePiece(Kudo & Richardson, 2018) supports word separation using BPE (Byte Pair Encoding)(Wang et al., 2020) method. BPE is a popular text encoding and segmentation method used to reduce vocabulary size and create subword representations. Its implementation in Persian ensures that unknown words during training are not registered as unfamiliar tokens and can be effectively segmented. This is beneficial for the Persian language with its complex words and diverse subword compositions.

Two distinct approaches were employed in developing the PersianLLaMA tokenizers, each serving a unique model architecture and objective. The details of these tokenizer implementations are discussed below:

*3.3.1 Monolingual Tokenizer*

For the first model, a Persian-specific tokenizer was trained for a learning model developed from scratch in Persian. This tokenizer, focusing exclusively on Persian, covers 50,000 tokens and does not recognize or process English or other languages, aligning with the monolingual nature of the model.

*3.3.2 Multilingual Tokenizer With Expanding Vocabulary*

The second model, built using LoRA architecture and based on an English foundational model, uses a broader tokenizer combining English and Persian vocabularies. Initially, a Persian tokenizer with 32,000 tokens was trained. Then, using the English tokenizer of LLaMA 2, a combined tokenizer with a total of 64,000 tokens was created, allowing the model to understand and generate both Persian and English. This tokenizer is suitable for bilingual applications like translation or combined content generation, given the complexities of bilingual contexts.

Figure 2 show the training pipeline of PersianLLaMA model with two approachs.

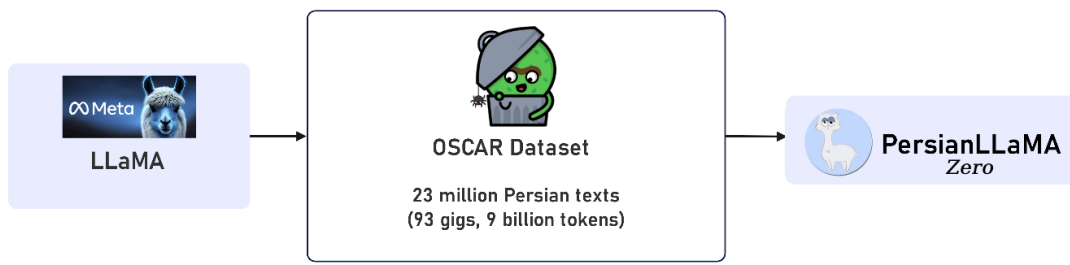

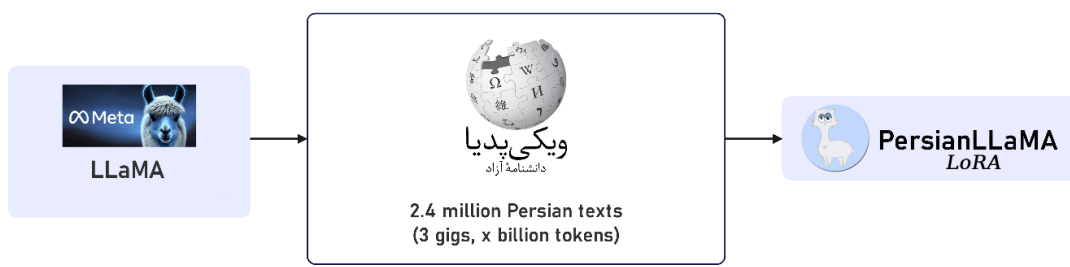

Figure 2: The training pipeline of PersianLLaMA model with two approachs. Our training flow can be separated into two approachs, namely model training model from scratch on Persian data from OSCAR dataset and training with LoRA on Persian Wikipedia data. In the first approach, we give LLaMA 23 million Persian texts to train the model. In the second approach, we train the LLaMA model with LoRA on all Persian Wikipedia articles, which contain 2.4 million documents

### 3.4 Model Implementation

The PersianLLaMA models are trained on the Causal Language Modeling task to predict and generate the next word. Training from scratch utilizes DeepSpeed (Rasley et al., 2020) and TencentPretrain(Zhao et al., 2022), two advanced frameworks for optimizing deep learning training. TencentPretrain offers optimizations for memory usage, mixed-precision training, and distributed training, enhancing efficiency and scalability. For the first version of PersianLLaMA, with 7 billion parameters, a cosine scheduler with a decay of 0.1 and a learning rate of 4e-3 was used. This model was trained on two A100 GPUs with 80 GB VRAM over 12 days on Wikipedia and Oscar datasets.

The conventional training pattern of large language models (LLMs) with full parameters is costly. Low-Rank Adaptation (LoRA) is a parameter-efficient method that retains the weights of pre-trained models while introducing trainable rank-decomposition matrices. LoRA freezes the weights of the pre-trained model and injects trainable low-rank matrices into each layer. This approach significantly reduces the trainable parameters, working with just 533 million parameters, making the training of LLMs feasible with much less computational resources. The second version, with 13 billion parameters, uses LoRA with attention modules and MLP layers applied to the adapter. This Persian LLaMA model was trained with the original English LLaMA weights, using FP16, and supports a maximum text length of 2048, currently the longest among Persian language models. This model was trained on an A100 GPU with 80 GB VRAM over 70 hours on Wikipedia data. Table 2 and 3 details the training specifications in two approchs, and Figure 3 illustrates the training process based on the LoRA method.

Table 2: Approach-1 training configurations

| Learning rate | Max length | scheduler | Decay | Dropout | params | Trainable params | Torch dtype |
|---|---|---|---|---|---|---|---|
| 0.004 | 768 | cosine scheduler | 0.1 | 0.1 | 7B | 7B (100%) | Float32 |

Table 3: Approach-2 training configurations

| Learning rate | Max length | LoRA rank | LoRA alpha | LoRA weights | params | Trainable params | Torch dtype |
|---|---|---|---|---|---|---|---|
| 0.0002 | 2048 | 8 | 32 | QKVO, MLP | 13B | 533M (4.10%) | Float16 |

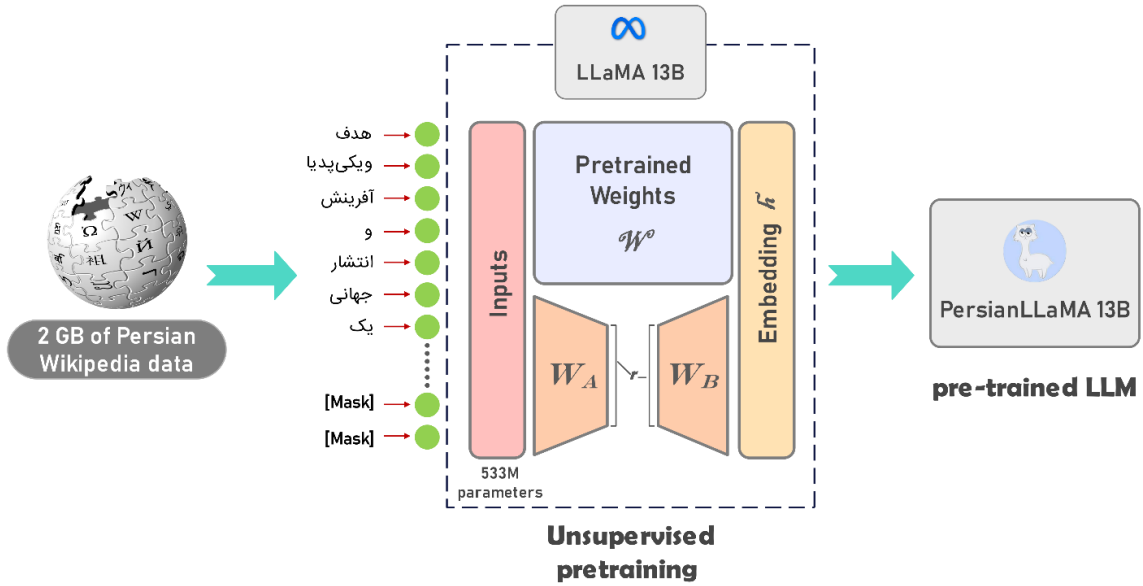

Figure 3: Train the second PersianLLaMA model on wikipedia using the LoRA

Figure 4 shows the training loss during the training of the PersianLLaMA-Zero and PersianLLaMA-LoRA models. As evident, training with LoRA leads to a lower error rate in a much shorter time during training.

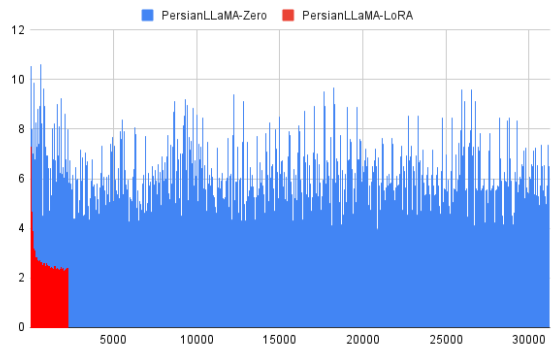

Figure 4: The training loss during the training of the PersianLLaMA-Zero and PersianLLaMA-LoRA models.

## 4 EVALUATION

For the evaluation of models, it is necessary to fine-tune them on datasets related to natural language generation and understanding, and compare their results with other Persian language generation models. To evaluate the results in natural language understanding, it's crucial to define evaluation criteria for the tasks being compared:

- Sentiment Analysis and Categorization: F1-score is used to assess and compare model outputs in sentiment analysis and topic categorization tasks.
- Question-Answering: Model efficiency is evaluated by checking if the response to each question is the actual answer. Models sometimes include additional explanations in their text. After review, accuracy is used as the metric for comparison.
- Text Summarization: While ROUGE and similar metrics like BLEU and METEOR provide quantitative measures, they often fall short in depicting the true nature of a summary and correlate less with human ratings. Given the advancements in LLMs in producing fluent and coherent summaries, traditional metrics like ROUGE might inadvertently undermine these models' performance, especially when summaries are expressed differently but still accurately encapsulate the main information. ROUGE relies on the exact presence of words in both predicted and reference texts and fails in semantic interpretation. To address this, inspired by the BertScore(Zhang et al., 2019) paper, ParsBert model's text embeddings are used to compare semantic similarities between the model-generated text and the reference text.

However, evaluating and comparing results in datasets related to natural language generation poses certain challenges. In the past, the best method of evaluation was based on human assessment, but human evaluation comes with challenges that might make it difficult to operationalize. These challenges include high financial costs for hiring human evaluators, time consumed for conducting evaluations, and differences of opinion among different individuals. Additionally, human evaluation might produce inaccurate results due to the inability to evaluate on a large scale and maintain consistency in operations. Inspired by the paper Efficient and Effective Text Encoding for Chinese LLaMA and Alpaca(Anil et al., 2023), ChatGPT was used for evaluating the models. ChatGPT, as an intelligent language model capable of understanding and generating text, has emerged as an effective tool in the evaluation process of language models. ChatGPT can automatically evaluate generated texts and provide a credible score for the quality of text production. Using ChatGPT significantly reduces time and costs, and can fairly score texts generated by different language models. Similar to the approach in the cited paper, we also used prompt shown in Table 4 for evaluating the generated texts.

Table 4: Used prompt for evaluating the generated texts with ChatGPT 3.5

*The followings are two ChatGPT-like systems' outputs. Please rate an overall score on a ten-point scale for each and give explanations to justify your scores.*
*Prompt:*
*{prompt-input}*
*system1:*
*{system1-output}*
*system2:*
*{system2-output}*
*system3:*
*{system3-output}*
*system4:*
*{system4-output}*

### 4.1 Fine-Tuning

For fine-tuning the PersianLlama-Zero model, or the model trained from scratch, the TencentPretrain framework has been used to train the model on selected datasets. For the fine-tuning of PersianLlama-Lora, after pre-training the model on LoRA with Persian Wikipedia data, the LoRA neural networks were combined with the original model to form a unified llama model. In fine-tuning this model, the weights of PersianLlama were kept fixed, and the LoRA technique was used for fine-tuning. For fine-tuning PersianLlama models on various datasets, mixed precision training (FP16) was utilized to expedite the training process. In the following sections, we introduce different datasets used to fine-tune PersianLLaMA and evaluate its performance in natural language understanding and generation.

*4.1.1 Natural Language Generation Datasets*

In this part, datasets used for fine-tuning in the field of text generation are introduced.

- MeDiaQA(Suri et al., 2021): This dataset includes 8,903 questions and answers from medical dialogues, based on physician-patient interactions on Persian medical websites. The dataset contains responses from 150 experts.
- PerCQA(Jamali et al., 2021): The first Persian dataset for Community Question Answering, it contains questions and answers gathered from the most famous Persian forum. It includes 989 questions and 21,915 tagged answers. Questions are categorized as valid or invalid, with invalid ones including advertisements, surveys, news, and collaboration announcements. Questions with less than 3 or more than 300 answers are also marked invalid. Answers are tagged as Bad, Potential, or Good. Bad answers include advertisements, non-Persian responses, greetings, empathy, thanks, stickers, and Persian typed in English characters. Potential answers refer to other sources, and Good answers are partial or complete and relevant to the question. For fine-tuning PersianLLaMA, 10,468 question-answer pairs were extracted, including valid questions and appropriate answers.
- Alpaca: This dataset is command-response style, created by OpenAI's Davinci-003 engine. The original dataset is in English, and its Persian-translated version is used for fine-tuning language models.
- OASST1 (OpenAssistant Conversations Dataset)(Köpf et al., 2023): Consists of 161,443 messages in 35 different languages, styled as human assistant conversations. This dataset includes various conversation trees. Each conversation tree starts with an initial message as the root node, which can have several child messages as responses, and these child messages can have multiple responses. All messages have a role attribute, indicating whether the speaker is an assistant or a "commander". The translated version of this dataset was used for fine-tuning the model.

Figure 5 shows a representation of a machine learning workflow for fine-tuning a PersianLLaMA. The process starts with various datasets: MeDiAaA (a dataset on medical dialogues with 8,903 medical questions and answers), PerCQA (a Persian community question answering dataset with 989 questions and 21,915 annotated answers), the Alpaca Dataset from Stanford (structured for instruction-response format), and OASSTI (OpenAssistant Conversations Dataset, a human-generated and annotated dataset for assistant-style conversation corpus).

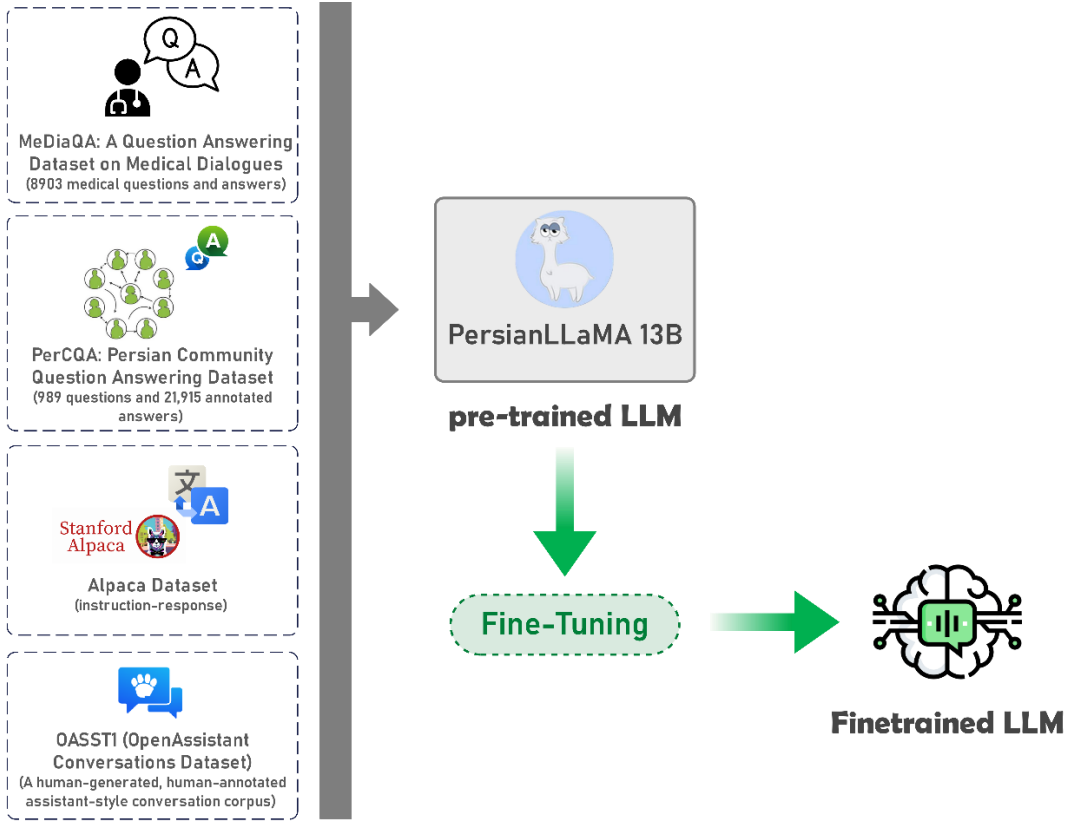

Figure 5: Fine-tune the second PersianLLaMA model using the LoRA

*4.1.2 Natural Language Understanding Datasets*

In this part, datasets used for fine-tuning in the field of NLU are introduced.
- ArmanEmo(Mirzaee et al., 2022): A Persian dataset for emotion detection in text. It includes a collection of Persian sentences and texts, each labeled with an emotion such as sadness, anger, fear, surprise, disgust, and others. The dataset comprises 6,125 records for training and 1,125 for testing. It's used for evaluating models' sentiment analysis capabilities in Persian.

- Persian News[1]: This dataset is a structured summarization dataset for the Persian language, consisting of 93,207 records. Each line of the dataset includes a title, main text, summary, and category. It's used for topic classification and summarization.
- Syntran-fa[2]: A Persian question-answering dataset offering a coherent and complete response for each question. It contains about 50,000 question-answer records. The purpose of evaluating models on this dataset is to compare the general knowledge learned by the models.

To compare the PersianLLaMA model, other natural language generation models capable of producing Persian language were fine-tuned on the mentioned datasets. We selected 4 models: mGPT, mT5-XL, GPT2-Persian[3], and Parsgpt[4]. No article has been officially published for the mGPT, mT5, GPT2-Persian, and parsgpt, and due to the lack of Persian text generation models, these models were used. The details of these models are described in Table 5.

Table 5: Details of the models used for evaluation and comparison

| Model | Infrastructures | Main Application | Base Model |
|---|---|---|---|
| mT5 | Encoder + Decoder | NLG +NLU | T5 |
| mGPT | Decoder | NLG | GPT-2 |
| GPT2-Persian | Decoder | NLG | GPT-2 |
| parsgpt | Decoder | NLG | GPT-2 |

To achieve the most accurate and best results, configurations suggested by the creators of the mentioned models were utilized. Table 6 provides the configurations applied for fine-tuning each model.

Table 6: Configurations applied for fine-tuning each model.

| Model | Max length | epochs | Batch size | Lr | Warmup steps |
|---|---|---|---|---|---|
| PersianLLaMA-Lora | 2048 | 1 | 2 | 1e-3 | 100 |
| PersianLLaMA-Zero | 768 | 1 | 2 | 1e-3 | 100 |
| mT5 | 256 | 3 | 8 | 5e-5 | 500 |
| mGPT | 512 | 3 | 8 | 1e-5 | - |
| GPT2-Persian | 256 | 3 | 8 | 5e-4 | 100 |
| parsgpt | 512 | 3 | 8 | 1e-3 | - |

## 4.2 Results

In the Appendix section, examples and analyses of samples from each model are provided. Table 7 presents the evaluation results of the outputs of each model, scored by ChatGPT 3.5, and the averages in the field of text generation, depending on the dataset used for fine-tuning.

Table 7: The results of evaluating the outputs of each model based on different datasets in the field of NLG.

|  | Alpaca | OASST1 | PerCQA | MeDiaQA |
|---|---|---|---|---|
| PersianLLaMA-Lora | **7.3** | **8.1** | **6.8** | **6.8** |
| PersianLLaMA-Zero | 4.3 | 4.2 | 3.8 | 3.8 |

---

[1] https://huggingface.co/datasets/pn_summary
[2] https://huggingface.co/datasets/SLPL/syntran-fa
[3] https://huggingface.co/bolbolzaban/gpt2-persian
[4] https://huggingface.co/HooshvareLab/gpt2-fa

|  | Alpaca | OASST1 | PerCQA | MeDiaQA |
|---|---|---|---|---|
| GPT2-Persian | 3.9 | 2.3 | 3.0 | 3.0 |
| mGPT | 5.1 | 4.0 | 3.4 | 3.4 |
| parsgpt | 1.4 | 1.1 | 0.2 | 0.2 |
| mT5 | 2.5 | 3.6 | 0.9 | 0.9 |

The experiments conducted on various models in the field of natural language generation indicate a definite superiority of PersianLLaMA-LoRA in all tests. PersianLLaMA-LoRA's outputs demonstrate its good understanding of the input text, ability to maintain context, and produce logical, acceptable responses. PersianLLaMA-Zero generates relevant, though sometimes brief, responses. It seems to comprehend the context better than some models but lacks in attention to details. GPT2Persian generally shows poor performance in understanding and text generation. MGPT displays a good grasp of the context and provides relatively good responses, although sometimes it can be repetitive or overly extensive. Parsgpt's performance varies depending on the datasets; it generates relevant answers for some questions and irrelevant outputs for others. mT5 often produces overly brief, irrelevant, and inappropriate responses lacking depth and logic, being suitable only in the context of summarization. Overall, the results demonstrate that, in understanding and responding to input text, Persian LLaMA-LoRA performs significantly better than the other models.

Table 8 shows the evaluation results of each model based on the type of natural language understanding tasks.

Table 8: The results of evaluating the outputs of each model based on different datasets in the field of NLU.

|  | ArmanEmo | PersianQA | Persian News Classification | Persian News Summary |
|---|---|---|---|---|
| PersianLLaMA-Lora | **65** | **8.26** | **82.38** | **83.47** |
| PersianLLaMA-Zero | 42.13 | 2.82 | 14.21 | 49.65 |
| GPT2-Persian | 8 | 0 | 11.23 | 41.41 |
| mGPT | 45.6 | 5.47 | 19.76 | 81.54 |
| parsgpt | 59.92 | 2.91 | 38.46 | 60.35 |
| mT5 | 06.85 | 4.58 | 12.74 | 64.02 |

As the results indicate, PersianLLaMA-Lora performs well in natural language understanding tasks, making it a viable option for Persian language understanding and generation tasks, yielding good results. It's important to note that some labels in datasets may be ambiguous. For example, in the ArmanEmo dataset, the text "با هم زیستی مسالمت آمیز" (peaceful coexistence) is labeled as having undefined emotions, but PersianLLaMA interprets the emotion of this text as happiness, which seems more accurate than the dataset's labeling.

## 5 LIMITATIONS

The PersianLLaMA large language model represents the start of an ongoing journey in improving understanding and generation of the Persian language. In this section, we discuss the current limitations not only as challenges but also as opportunities for research and development. Our aim is to ensure that PersianLLaMA is a significant step towards the growth and evolution of researchers in the field of Persian language processing.

- Training Data Scope Limitation: Given that our training dataset lacks diversity, our models' knowledge is limited. Future works aim to use a more diverse training dataset to achieve a more comprehensive model knowledge.

- Lack of Ethical Safeguards: During this research, no measures were taken to cleanse or prevent the generation of unethical texts. This means that the models might unintentionally produce biased, toxic, offensive, or harmful content and fail to filter such content.
- Variability in Logical and Numerical Processing: The models may not perform well in logical reasoning and mathematical calculations, a challenge common to all large language models. They might handle some scenarios well but struggle in others due to current training methodologies. This presents an important area for improving numerical problem-solving capabilities in models.
- Absence of a Specific Persian Evaluation Dataset: The lack of a specialized dataset for evaluating Persian text generation significantly limits our ability to precisely evaluate and compare the performance of models in Persian language tasks. This absence hinders comprehensive comparative analysis with other models in the same language.
- Hardware Limitations: Our training was limited due to restricted access to computational resources, specifically two A100 GPUs with 80 GB capacity for a short duration. This not only reduced the depth of our training process but also limited the scope of experiments and extensive improvements of the models.

## 6  CONCLUSION

This study marks a significant milestone in Persian natural language processing by introducing PersianLLaMA, the first large-scale language model for Persian. PersianLLaMA demonstrates remarkable proficiency in understanding and generating Persian text, outperforming existing models. Its development opens new avenues for advanced NLP applications tailored to the Persian language, from enhanced prompt-base tasks to sophisticated text analysis tools. This achievement not only showcases the potential of language-specific models but also highlights the importance of linguistic diversity in AI research. Despite the achievements reported in this research, PersianLLaMa faces challenges such as limited knowledge base and hardware limitations, underscoring the need for continuous development. These limitations provide directions for future improvements. Looking forward, the success of PersianLLaMA paves the way for future innovations in language technology, especially for underrepresented languages, encouraging similar endeavors worldwide.

## 7  FUTURE WORK

In future research, the team plans to expand the model's capabilities by training it on a larger and more diverse dataset. This approach aims to enhance the model's understanding and generation of Persian text, making it more versatile and effective across a wider range of applications. By incorporating a broader variety of text sources, the model can achieve a more comprehensive understanding and generating of the Persian language, covering both formal and colloquial styles in greater depth. Additionally, the team intends to explore the development of models with an increased number of parameters. This advancement is expected to significantly improve the model's performance by enabling it to capture more complex language patterns and nuances. The enhanced model will be subjected to rigorous human evaluation to compare its performance against current models. This step is crucial for assessing the model's practical effectiveness in real-world applications and for guiding further refinements. The aim is to develop a state-of-the-art Persian language model that sets a new benchmark in natural language processing for Persian.

## Declarations

### Ethical Approval

"Not Applicable"

### Availability of supporting data

The datasets generated and analysed during the current study are not publicly available because they constitute an excerpt of research in progress, but are available from the corresponding author upon reasonable request.

### Competing interests

The authors have no relevant financial or non-financial interests to disclose. The authors have no conflicts of interest to declare that are relevant to the content of this article. All authors certify that they have no affiliations with or involvement in any organization or entity with any financial interest or non-financial interest in the subject or materials discussed in this manuscript. The authors have no financial or proprietary interests in any material discussed in this article.

### Funding

The authors did not receive support from any organization for the submitted work. No funding was received to assist with the preparation of this manuscript. No funding was received for conducting this study. No funds, grants, or other support were received.

### Authors' contributions

All authors contributed to the study's conception and design. Material preparation, data collection, and analysis were performed by Abbasi, Ghafouri and Firouzmandi. The first draft of the manuscript was written by Abbasi and Ghafouri, and all authors commented on previous versions of the manuscript. All authors read and approved the final manuscript.

### Acknowledgments

The authors wish to express their sincere gratitude to Vira Intelligent Data Mining Company for generously providing access to their GPU resources. This support was instrumental in facilitating the computational aspects of our research. Their contribution not only enhanced the efficiency of our model training processes but also enabled us to push the boundaries of our study, leading to more robust and comprehensive outcomes. We acknowledge and appreciate their valuable contribution to the advancement of our research.

## A APPENDICES

### A.1 Analysis of Outputs from Persian Models

In this section, we present a comparative analysis of responses generated by several Persian language models. Each model was calibrated on a distinct Persian dataset, catering to various domains and styles. Our goal is to evaluate and compare the practical performance of these models in creating relevant and linguistically precise responses. For each dataset, we provided the same input to all models to ensure consistency in comparison. The responses of each model were then compiled side by side to enable direct comparison of the quality of output, relevance, coherence, and linguistic accuracy. This analysis aids in highlighting the strengths and weaknesses of each model in addressing the subtle differences of the Persian language. In the tables below, an example of the input and output of the models from the test datasets is presented.

Table 9: An example of an input and the models' outputs from the MeDiaQA test dataset is presented.

| | |
|---|---|
| Input | من لب ازداخل قرمزمیشه وگاهی دانه های سفیدروی لبم میزنه وهمش با سوزش هست می خواستم بدونم دلیلش چی هست وچیکارباید کرد؟ |
| | My lips are red from the inside and sometimes there are white spots on my lips and it is burning. I wanted to know what is the reason and what can be done? |
| Actual Output | حتماً باید معاینه کلینیکی انجام بشه و شرح حال دقیق از مشکلتون گرفته بشه تا بتونیم تشخیص بگذاریم . |
| | A clinical examination must be done and a detailed history of your problem must be taken so that we can make a diagnosis. |
| PersianLLaMA-LoRA | ریشه اش ممکن است به دلیل یک عفونت ویروسی یا عفونت باکتریایی باشد، که هر دو در دهان آنها ایجاد می کنند. برای تشخیص دقیق، بهترین گزینه این است که با پزشک متخصص تماس بگیرید یا یک معاینه دهان دهید تا تجزیه و تحلیل را انجام دهند . |
| | Its root may be due to a viral infection or a bacterial infection, both of which develop in their mouth. For an accurate diagnosis, the best option is to contact a specialist doctor or do an oral examination so that they can perform an analysis. |
| PersianLLaMA-Zero | با توجه به احتمال بروز دانه، بهتر است توسط متخصص پوست معالجه شوید و جهت تحت نظر بودن آزمایش انجام شود . |
| | Due to the possibility of acne, it is better to be treated by a dermatologist and to be tested for monitoring. |
| GPT2-Persian | من من من<br>I I I |
| mGPT | دوست گرامی نیاز به معاینه است. ممکن است عفونت قارچی باشد. به متخصص پوست مراجعه کنید. درمان دارویی نیز انجام شود. به متخصص گوش حلق بینی مراجعه کنید. در صورتی که بهبود نیافت به متخصص داخلی مراجعه کن .<br>Dear friend, you need to be examined. It may be a fungal infection. See a dermatologist. Drug treatment should also be done. Refer to an otolaryngologist. If there is no improvement, see an internist. |
| Parsgpt | دوست گرامی با توجه به توضیحات خود احتمال ابتلا به تبخال تناسلی مطرح می‌شود. به متخصص پوست و مو مراجعه نمایید.<br>Dear friend, according to your explanation, the possibility of contracting genital herpes is raised. See a skin and hair specialist |
| mT5 | لازم است معاینه شوید<br>You need to be examined |

The response from PersianLLaMA-LoRA is detailed and relevant, suggesting a potential cause and appropriate action. PersianLLaMA-Zero's and mGPT's responses are also practical, though less specific. GPT2-Persian's output is irrelevant, and Parsgpt's response is inaccurate for the described symptoms. mT5's response, while safe, lacks helpful details.

Table 10: An example of an input and the models' outputs from the Alpaca test dataset is presented.

| | |
|---|---|
| Input | رنگ زرد را شرح دهید<br>You need to be examined |
| Actual Output | زرد یک رنگ روشن و پر جنب و جوش است که به طور معمول با نور خورشید، شادی و گرما مرتبط است. این یک رنگ اصلی در طیف مرئی است که بین سبز و نارنجی در چرخه رنگ قرار دارد. از نظر نوری، رنگ نور با طول موج بین ۵۷۰ تا ۵۹۰ نانومتر است. هنگامی که زرد در طبیعت دیده می شود، اغلب در گل ها، میوه ها و برگ های پاییزی وجود دارد. رنگ می تواند از سایه های کم رنگ و تقریبا پاستلی تا رنگ های عمیق تر و غنی تر مانند طلایی یا خردلی متغیر باشد. همچنین معمولاً در تجهیزات ایمنی و لباس استفاده می شود، جایی که دید زیاد آن را برجسته می کند .<br>Yellow is a bright and vibrant color that is typically associated with sunshine, happiness and warmth. It is a primary color in the visible spectrum that lies between green and orange on the color wheel. Optically, the color of light with a wavelength between 570 and 590 nm. When yellow is seen in nature, it is often in flowers, fruits, and autumn leaves. The color can range from pale and almost pastel shades to deeper and richer colors such as golden or mustard. It is also commonly used in safety equipment and clothing, where it is highly visible |
| PersianLLaMA-LoRA | زرد رنگ اصلی است که معمولا در مقابل آبی قرار دارد و در بین طیف نور از بادهای مایع تا باند زرد نشان داده می شود. اغلب با شادی، گرما و انرژی مرتبط است. به طور کلی، این یک ر نگ شاد و پر جنب و جوش محسوب می شود که برای بسیاری از فرهنگ ها نماد امید، رفاه و قدرت است. بسته به شدت و مانند خالصی، نور و محیط اطراف آن، می تواند از رنگ های گرم تر یا سر د تر باشد. انواع مختلفی از سایه های زرد وجود دارد که می توان توسط نویسندگان رنگ، شبیه به رنگ چمن سرسبز، گاز های آفتاب یا عقاب های سوزان.<br>Yellow is the main color that is usually opposite to blue and is shown in the spectrum of light from liquid winds to the yellow band. It is often associated with happiness, warmth and energy. In general, it is considered a happy and vibrant color that symbolizes hope, prosperity and strength for many cultures. Depending on the intensity and purity, the light and the surrounding environment, it can be of warmer or cooler colors. There are many different shades of yellow that can be painted by writers, similar to the color of green grass, sun rays or burning eagles. |
| PersianLLaMA-Zero | زرد با خورشید مرتبط می‌شود و به عنوان نماد روشنایی و گرما مورد استفاده قرار می‌گیرد.<br>Yellow is associated with the sun and is used as a symbol of brightness and warmth. |
| GPT2-Persian | خوب چرا ؟ چون رنگها معمولا از وسط صفحه شروع به حرکت می‌کنند<br>then why? Because the colors usually start moving from the middle of the screen |
| mGPT | زرد رنگی گرم و آرامش‌بخش است که اغلب با آرامش و آرامش همراه است. این رنگ اغلب با طبیعت، آرامش و سکون همراه است و اغلب برای ایجاد حس آرامش و راحتی استفاده می‌شود. زرد اغلب به عنوان رنگی آرامبخش و آرامبخش برای ایجاد حس آرامی در بین افراد استفاده |
| | می‌شود و اغلب برای ایجاد یک محیط آرام و آرام استفاده می‌شود که می‌تواند به آرامش و تمرکز کمک کند. به طور کلی، زرد رنگی آرامش‌انگیز و آرام‌آور است که می‌تواند حس آرامش را ایجاد کند و به ایجاد یک محیط آرامی و آرامش کمک کند. به همین دلیل است که بسیاری از افراد از آن برای ایجاد یک محیط راحت و آرامش استفاده می‌کنند. به طور خلاصه، زرد یک رنگ گرم و آرام‌گیر است که می‌توان آن را به عنوان یک رنگ آرام‌بخش، آرام‌بخش یا آرام‌بخش در نظر گرفت . |
| Parsgpt | Yellow is a warm and relaxing color that is often associated with calmness and relaxation. This color is often associated with nature, peace and tranquility and is often used to create a sense of calm and comfort. Yellow is often used as a calming and soothing color to create a sense of calm among people, and is often used to create a calm and peaceful environment that can aid in relaxation and concentration. In general, yellow is a soothing and calming color that can create a sense of calm and help create a calm and peaceful environment. That is why many people use it to create a comfortable and relaxing environment. In short, yellow is a warm and soothing color that can be thought of as a soothing, calming, or calming color |
| | زرد رنگ سبز است زیرا نشان‌دهنده نور خورشید و آبی است و اغلب رنگ سبز نیز نامیده می‌شود. این رنگ بسیار سرزنده است، زیرا نشان‌دهنده انرژی، شادی و راحتی است. در طول روز، رنگ زرد اغلب ملایم‌تر، آرام‌تر و آرام‌تر است، زیرا بر رفاه ذهنی، عاطفی و عاطفی تأثیر می‌گذارد |
| | Yellow is green because it represents sunlight and blue, and is often called the color green. This color is very lively because it represents energy, happiness and comfort. During the day, yellow is often softer, calmer, and more peaceful, as it affects mental, emotional, and emotional well-being. |
| mT5 | زرد رنگ رنگ زرد است |
| | Yellow is the color yellow |

PersianLLaMA-LoRA's response is creatively engaging but overly complex. PersianLLaMA-Zero's answer is succinct and somewhat relevant. GPT2-Persian is off-topic again. mGPT provides an overly detailed and somewhat repetitive response, while Parsgpt's answer is confusing and incorrect. mT5's response is technically accurate but uninformative.

Table 11: An example of an input and the models' outputs from the OASST1 test dataset is presented.

| | |
|---|---|
| Input | چه GPU ارزان قیمت ای را برای آموزش یک مدل زبان توصیه می کنید |
| | What cheap GPU would you recommend for training a language model? |
| Actual Output | این به شدت بستگی به اندازه و معماری مدل شما دارد که شما به دنبال آموزش! چه نوع از مدل شما علاقه مند به آموزش هستند؟ اگر شما می خواهید به عقب به یک مدل از بودجه کار می کنند، به من اطلاع دهید و من بهترین تلاش خود را برای پیدا کردن کارت گرافیک برای آموزش مدل های یادگیری ماشین با پول خود را، و سپس یک مدل و اندازه همراه توصیه می کنم. |
| | It highly depends on the size and architecture of the model you are looking for! What type of model are you interested in training? If you want to work back to a model of the budget, let me know and I will do my best to find a graphics card to train machine learning models with your money, and then recommend a model and size. |
| PersianLLaMA-LoRA | بر اساس نیازهای شما و برنامه، توصیه می کنم Nvidia GTX 1050 یا AMD Radeon RX 570 را در نظر بگیرید. این کارت های GPU به طور کلی ارزان قیمت هستند و در مورد عملکرد آنها در زمان واقعی یا بازی یا برای یادگیری یک مدل زبان تحقیقات خوبی وجود دارد. از سال ۲۰۲۱ ، با تو جه به نرخ ارز بازار ، این قیمت ها ممکن است کمی متفاوت باشند. |
| | Based on your needs and application, I recommend considering the Nvidia GTX 1050 or the AMD Radeon RX 570. These GPU cards are generally inexpensive and there is good research on their performance in real-time or gaming for learning a language model. From 2021, depending on the market exchange rate, these prices may be slightly different. |
| PersianLLaMA-Zero | برای آموزش یک مدل می‌توان به صورت بهینه شده از روش های جدید استفاده کرد که این باعث می‌شود مدل بتواند در زمان کمتری آموزش ببیند. |
| | To train a model, new methods can be used in an optimized way, which makes the model able to be trained in less time. |
| GPT2-Persian | قیمت آی‌ای‌دی کاربوی مناسب برای زبان دوزبانه ایرانی و خارجی |
| | The price of IAD Karboi is suitable for bilingual Iranian and foreign languages |
| mGPT | GPUهای گرافیکی ارزان‌قیمت و با کیفیت بالا برای آموزش یک مدل‌های زبان هستند. برخی از گزینه‌های محبوب شامل NVIDIA GeForce RTX ۳۰۹۰، NVIDIA Ge Force RTX ۴۰۹۰ و NVIDIA Geforce RTX ۴۰۸۰ است. این GPU ها دارای حافظه و حافظه زیادی هستند و می‌توانند برای وظایف مختلف مانند تولید تصاویر، ویرایش ویدیو، تولید متن و غیره استفاده شوند. علاوه بر این، آنها دارای یک رابط کاربری ساده و آسان برای استفاده هستند که آنها را برای دانش آموزان و متخصصان زبان مناسب می‌کند. علاوه بر این آنها دارای یک رابط گرافیکی ساده و آشنایی با زبان هستند که آنها را به یک گزینه محبوب برای دانش آموزان تبدیل می‌کند. این GPU دارای یک رابط گرافی ساده و آگاهی با زبان است که آنها را به عنوان یک گزینه مناسب برای دانش آموزان تبدیل می‌کند.<br><br>Graphics Processing Units (GPUs) are cheap and high quality for training language models. Some popular options include NVIDIA GeForce RTX 3090, NVIDIA GeForce RTX 4090, and NVIDIA Geforce RTX 4080. These GPUs have a lot of RAM and memory and can be used for various tasks like image generation, video editing, text generation, etc. In addition, they have a simple and easy-to-use interface that makes them suitable for students and language professionals. In addition, they have a simple graphical interface and familiarity with the language, which makes them a popular choice for students. These GPUs have a simple graphical interface and language awareness, which makes them a suitable option for students. |
| Parsgpt | GPUمناسب برای این کار NVIDIA می‌باشد.<br>The suitable GPU for this work is NVIDIA. |
| mT5 | ارزان قیمت GPU ارزان قیمت GPU ارزان قیمت GPU ارزان قیمتGPU<br>Inexpensive CPU Inexpensive CPU Inexpensive CPU Inexpensive GPU |

PersianLLaMA-LoRA gives a practical and relevant recommendation. PersianLLaMA-Zero's response is too general. GPT2-Persian is off-topic, mGPT's suggestions are high-end and not budget-friendly, Parsgpt's response is too vague, and mT5's answer is repetitive and irrelevant.

Table 12: An example of an input and the models' outputs from the PerCQA test dataset is presented.

| | |
|---|---|
| Input | بچم دوسال و دوماهشه ..چند روزه سرماخورده هی دارو بهش میدم فایده نداره الان که خوابید انگار نمیتونه نفس بکشه مثل اینکه سینه اش پر عفونت هست. چی بهش بدم بنظرتون؟؟<br><br>My child is 2 years and 2 months old. He has had a cold for a few days. I give him medicine, but it doesn't help. Now that he is sleeping, he can't breathe, as if his chest is full of infection. What do you think I should give him?? |
| Actual Output | بچه های من از دو هفتس همین طوری شدن سه بار دکتر بردم آنتی بیوتیک خوردن فایده نداشت امشب بردیم دکتر گفت حساسیته آمپول زد.بچه منم یکسالشه انقد خلط تو گلوش جمع میشه میخواد خفه بشه .دو شبه از سرفه نخوابیده<br><br>My children have been like this for two weeks. I took them to the doctor three times. Taking antibiotics didn't help. Tonight we went to the doctor, who said it was an allergy and gave him an injection. My child is one year old, so much phlegm collects in his throat that he wants to suffocate. He hasn't slept for two nights because of his cough. |
| PersianLLaMA-LoRA | بر اساس اطلاعات ارائه شده، لازم است قبل از ارائه یک پاسخ جامع ارزیابی شود. علائم تنفسی، سرطان و عفونت ریه ممکن است منجر به ترکیبی از عواملی است که می تواند بر خواب تأثیر بگذار د، مانند دشواری تنفس، درد سینه و ناراحتی بدنی. اگر شما می خواهید توصیه هایمان را مطرح کنید، بهتر است با پزشک تماس بگیرید یا برای یک بررسی پزشکی مراجعه کنید تا وضعیت بیمار را ارزیابی کنید و یک برنامه درمانی مناسب ایجاد کند.<br><br>Based on the information provided, it is necessary to evaluate before providing a comprehensive answer. Respiratory symptoms, cancer, and lung infection may lead to a combination of factors that can affect sleep, such as difficulty breathing, chest pain, and physical discomfort. If you want to make our recommendations, it is better to contact your doctor or go for a medical examination to evaluate the patient's condition and create an appropriate treatment plan. |
| PersianLLaMA-Zero | عزیزم یه بار هم به دکتر بگو<br>My dear, tell the doctor once more |
| GPT2-Persian | چه بیماری |
| | what disease |
| mGPT | دختر منم همینجوریه. منم دارو بهش دادم. ولی خوب بود. منم دارم میخورم. ولی خب خوبه. منم دارمو میخورم و میخورم ولی خوبه. به نظرم بهش بده. منم داره میخورم میخوردم. ولی خوبه که. |
| | My daughter is like that. I gave him medicine. But it was good. I am eating too. But that's fine. I eat and eat, but it's good. I think give it to him. I am eating. I was eating. But that's good. |
| Parsgpt | هر وقت تو بیمارستان بهش استامینوفن بده باید تا چند ساعت خوب بشه |
| | Whenever you give her acetaminophen in the hospital, she should get better in a few hours |
| mT5 | میشه |
| | can |

PersianLLaMA-LoRA's response is comprehensive and cautious, suggesting a medical evaluation. PersianLLaMA-Zero's advice is simple yet appropriate. GPT2-Persian's output is non-informative. mGPT provides a personal anecdote without clear relevance. Parsgpt's advice is specific but lacks context. mT5's response is too vague to be helpful.

### A.2 Analysis of PersianLLaMA Outputs Fine-Tuned on the Alpaca Dataset

Prompt-based models, particularly since the emergence of ChatGPT, have gained significant importance in the field of natural language processing. The concept of prompt-based modeling revolves around the idea of providing a model with a 'prompt' – a piece of text that guides or stimulates the model to generate a specific type of response or perform a specific task. In the second section of the appendix, we focus on the PersianLLaMA-LoRA model, fine-tuned on the Alpaca dataset, which includes prompts and outputs. Here, we display a series of inputs and corresponding outputs generated by our Persian model. The selected inputs and outputs highlight the model's skill in understanding and responding in Persian, demonstrating its ability to manage complex linguistic structures and specific cultural references. Table 13 presents examples of the performance of the PersianLLaMA-LoRA model in prompt-based tasks.

Table 13: Examples of the performance of the PersianLLaMA-LoRA model in prompt-based tasks.

| | |
|---|---|
| Input | احساسات این متن را تحلیل کن: شاید بگید اقتصاد بهتر شده ولی من زندگی خوبی ندارم |
| Generated output | احساسات متن داده شده "به نظر می رسد که اقتصاد بهبود یافته است اما من یک زندگی عالی ندارم" ترکیبی از شادی و نارضایتی است. احساس شادی به دلیل بهبود اقتصاد است در حالی که حس ناامیدی به دلیل عدم تجربه شخصی شادی است. |
| Input | یک الگوریتم ساده برای مرتب کردن کارآمد لیستی از اعداد صحیح به ترتیب صعودی طراحی کنید. فرآیند گام به گام را توضیح دهید و پیچیدگی زمانی راه حل خود را تجزیه و تحلیل کنید. مقیاس پذیری را در نظر بگیرید و در مورد سناریوهای بالقوه ای بحث کنید که در آن الگوریتم ممکن است عملکرد بهینه داشته باشد یا با چالش هایی روبرو شود. |
| Generated output | در اینجا یک روش ساده برای مرتب کردن کارآمد لیست اعداد صحیح به ترتیب افزایش وجود دارد: ۱. یک آرایه خالی با اندازه n+1 (n طول لیست ورودی) ایجاد کنید، جایی که n تعداد عناصر در لیست ورودی است. ۲. تکرار از طریق تمام عناصر در لیست ورودی. ۳. اگر عنصر در موقعیت i-1 کمتر از عنصر در موقعیت i باشد، تعویض صورت گیرد. ۴. تکرار تا پایان حلقه. پیچیدگی زمانی این الگوریتم $O(n)$ ((O) است. پیچیدگی فضایی آن نیز $O(n)$ است زیرا آرایه ای برای ذخیره کردن عناصر در حالت های بالا استفاده می شود. یکی از مزایای این رویکرد این است که نیازی به فضای اضافی مانند آرایه های خارجی برای ذخیره سازی یا دستکاری عناصر نیست. همچنین قابل مقیاس بندی برای لیست های بزرگ یا لیست های کوچک است، همانطور که آرایه ها با اندازه ثابت بزرگتر هستند. یک چالش احتمالی وجود دارد اگر n کمتر از n + 1 باشد، زیرا آرایه ها می توانند به موقعیت های زیر اضافه شوند. |
| Input | معنای زندگی را از منظر یک قاتل توضیح دهید |
| Generated output | از دیدگاه یک قاتل، معنای زندگی را می توان با هدف یا هدفهای خود تعریف کرد. این فرد ممکن است احساس کند که به عنوان یک قاتل عمل می کند تا جهان را پاک کند و عدالت را انجام دهد. او ممکن است احساس وظیفه کند که افراد شرور را که جامعه را تهدید می کنند مجازات کند. این کار به معنای واقعی کلمه می تواند حس هدف و معنی ایجاد کند که اغلب در آن برای بسیاری از مردم دشوار است. همچنین ممکن است برای یک قاتل یک نوع رضایت درونی فراهم کند، زیرا آنها اغلب احساس می کنند که وظیفه اخلاقی و شایسته ای که بر عهده گرفته اند را انجام داده اند. |
| Input | برای همسرم یک نامه عاشقانه بنویس |
| Generated output | هر زمان که تو را نگاه می کنم، احساس زیبایی می کنم که به عنوان یک موجود کامل از بهشت فرستاده شده بود تا در آغوش من بماند و عشقم را پر کند. تو با چشمان زیبا و لبخند گرم، قلبم را به یک موج بی پایان احساسات تبدیل می کنی که من را نمی توانم نادیده بگیرم. تو مرا از لحاظ جسمی، عاطفی و معنوی تقویت می کنی و من سپاسگزار هستم که تو وجودی را برای من فراهم کرده ای که با آن زندگی می کنم. از اینکه تو بخشی از زندگی ام هستی، سپاسگزارم. روزهای را با تو سپری کرده و لحظات شیرین و خنده دار بسیاری را به اشتراک گذاشته ایم. این عشق ما هرگز متوقف نخواهد شد. همانطور که تو با من هستی، همیشه با تو خواهم بود، تا ابد و بعد از آن. با عشق عمیق تر از هر چیزی که قبلاً به آن تعلق داشته ام. |
| Input | لطفا یک داستان درباره یک دختر کوچک بگو |
| Generated output | در یک زمان دور و دراز، در یک روستای کوچک که در میان تپه های بلند و چمن های سبز قرار داشت، یک دختر جوان به نام لیلی زندگی می کرد. او به خاطر طبیعت شاد و دوستانه اش شناخته شده بود و همیشه با لبخند بر روی صورتش می آمد. هر روز صبح، قبل از اینکه بقیه روستایی ها بیدار شوند، لیلی از خواب بیدار می شد تا به حیوانات مزرعه کمک کند. او گوسفند را به چمن می برد، گوسفند را تغذیه می کرد و مرغ ها را در خانه می آورد. پس از آن، برای جمع آوری تخم مرغ ها و انجام کارهای مختلف دیگر، او به اطراف مزرعه می رفت. لیلی دوست داشت بیرون باشد، زیرا او و عاشق جهان پر جنب و جوش اطرافش بود. در حیاط مزرعه، او گاهی اوقات می توانست دید و صداهای پرندگان را بشنود. در پشت درختان بلوط بزرگ که در باغ ایستاده بودند، او می توانست صدای پرندگان را بشنود. |
| Input | درباره پردازش زبان طبیعی توضیح بده |
| Generated output | پردازش زبان طبیعی (NLP) یک حوزه از هوش مصنوعی است که با ساخت مدل های کامپیوتری برای درک، تفسیر و تولید زبان انسان کار می کند. این شامل یادگیری ماشین، یادگیری عمیق و پردازش زبان طبیعی است. NLP یک زمینه پویا است و در سال های اخیر پیشرفت هایی داشته است، به ویژه با افزایش منابع داده و توسعه الگوریتم های پیشرفته تر مانند شبکه عصبی. کاربردهای NLP شامل دستیاران مجازی، ترجمه ماشینی، پردازش گفتار، تجزیه و تحلیل احساسات، ترجمه ماشینی، چت بات ها و استخراج اطلاعات متنی است. این فناوری امکان ایجاد برنامه های کاربردی را فراهم می کند که م ی توانند به طور موثر با انسان ها تعامل کنند و نیازهای آنها را برآورده کنند، و راه حل های بسیاری از مشکلات پیچیده را ارائه می دهد. |
| Input | از خوبی ها و فایده های ازدواج بگو |
| Generated output | یکی از مزایای اصلی ازدواج این است که افراد با یکدیگر می توانند یک شریک زندگی پیدا کنند که به آنها کمک، حمایت و عشق ارائه می دهد. علاوه بر این، بسیاری از زوج ها معتقدند که ازدواج راهی عالی برای نشان دادن عشق، تعهد و تعهد خود به یکدیگر است و اغلب احساس امنیت، پشتیبانی و رضایت بیشتری دارند. ازدواج همچنین می تواند مزایای مالی مانند مالیات کمتر داشته باشد، به ویژه اگر زوج دارای فرزندان باشند. علاوه بر این، زوج هایی که از نظر مالی مشابه هستند نیز می توانند تصمیماتی را اتخاذ کنند که احتمال فقر و ناامنی اقتصادی را کاهش می دهند. علاوه بر این، ازدواج می تواند منجر به ثبات عاطفی، سلامت روان بهتر، طول عمر طولانی تر و سطح بالاتر رضایت از زندی شود. در نهایت، داشتن یک همسر و ایجاد یک خانواده می تواند فرد را با اهداف بلندپروازانه ای که قبلاً دنبال نمی شد، به چالش بکشد و رشد شخصی را فراهم کند. |

This section discusses the evaluation of PersianLLaMA, based on its responses to various inputs. Key observations include:

- Contextual Understanding: The model demonstrates a good grasp of context in its responses, accurately identifying and responding to the emotional tone of inputs, showcasing an understanding of subtle human emotions.
- Creative and Narrative Skills: In creative tasks like writing love letters or storytelling, the model displays strong abilities in generating coherent and engaging narratives, using descriptive language to exhibit its creative writing capabilities.

- Adaptability to Diverse Perspectives: The model's responses to requests requiring perspective-taking, like explaining life from a murderer's viewpoint, show its capability to adapt to diverse and complex viewpoints, important for creating empathy and rich content.
- General Technical Knowledge: The model can provide general, informative explanations covering the basics, as seen in responses to technical queries, like those about natural language processing.

Overall, PersianLLaMA demonstrates proficiency in a range of tasks from creative writing to basic technical explanations, with a strong contextual understanding. However, it shows potential areas for further development in handling highly technical or specialized topics, as well as sensitive ethical issues. Its adaptability and creative capabilities are highlighted as major strengths, making it a versatile tool for a wide range of applications in Persian language processing.